\documentclass[fleqn,10pt]{wlscirep}
\usepackage[utf8]{inputenc}
\usepackage[T1]{fontenc}
\title{Pre-training transformer-based framework on large-scale pediatric claims data for downstream population-specific tasks}

\author[1]{Xianlong Zeng}
\author[2]{Simon Lin}
\author[1,*]{Chang Liu}
\affil[1]{Ohio University, Electrical Engineering and Computer Science, Athens Ohio, 45701, USA}
\affil[2]{Nationwide Children's Hospital, Research Information Solutions and Innovation, Columbus Ohio, 43205, USA}

\affil[*]{liuc@ohio.edu}

\begin{abstract}

The adoption of electronic health records (EHR) has become universal during the past decade, which has afforded in-depth data-based research. By learning from the large amount of healthcare data, various data-driven models have been built to predict future events for different medical tasks, such as auto diagnosis and heart-attack prediction. Although EHR is abundant, the population that satisfies specific criteria for learning population-specific tasks is scarce, making it challenging to train data-hungry deep learning models. This study presents the Claim Pre-Training (Claim-PT) framework, a generic pre-training model that first trains on the entire pediatric  claims dataset, followed by a discriminative fine-tuning on each population-specific task. The semantic meaning of medical events can be captured in the pre-training stage, and the effective knowledge transfer is completed through the task-aware fine-tuning stage. The fine-tuning process requires minimal parameter modification without changing the model architecture, which mitigates the data scarcity issue and helps train the deep learning model adequately on small patient cohorts. We conducted experiments on a real-world claims dataset with more than one million patient records. Experimental results on two downstream tasks demonstrated the effectiveness of our method: our general task-agnostic pre-training framework outperformed tailored task-specific models, achieving more than 10\% higher in model performance as compared to baselines. In addition, our framework showed a great generalizability potential to transfer learned knowledge from one institution to another, paving the way for future healthcare model pre-training across institutions.

\end{abstract}
\begin{document}

\flushbottom
\maketitle
%
%
\thispagestyle{empty}


\section*{Introduction}
The rapid growth of electronic health records (EHR) promotes the use of data-driven modeling to improve care delivery and care management for individual patients. Specifically, novel applications are emerging that use state-of-the-art deep learning approaches such as recurrent neural networks~\cite{choi2017using}, convolutional neural networks~\cite{landi2020deep} and transformers~\cite{zeng2019multilevel}, to predict future medical events. With sufficient patient samples, these deep learning models can progressively extract relevant features and show promising model performance for various predictive tasks. For example, doctorAI, which yields promising results for predicting subsequent medical visits’ diagnosis and medication, was trained on a dataset with more than 200,000 patient records.

One prerequisite for training most deep learning models is the availability of substantial amounts of high-quality data~\cite{sun2017revisiting},~\cite{hedderich2018training}. Although the large-scale EHR database contains millions of patient records, these records are often not entirely applicable for various reasons, such as a limited number of cases for rare conditions and diseases~\cite{haines2020testing}, restricted access to the entire database due to privacy concerns~\cite{choi2017generating}, difficulty in data cleaning and merging~\cite{helgheim2019merging} (especially if collected from different institutions~\cite{seneviratne2018merging}). These limitations hinder the data acquisition process and, therefore, restrict the opportunities to develop data-hungry deep learning models, which may slow down the computational advances in healthcare and impede the improvement in care delivery.

Various methods have been proposed to address the data insufficiency issue, including synthetic data generation and incorporating medical domain knowledge. For example, Lee et al.~\cite{lee2020generating} proposed an autoencoder-based deep generative model to learn and synthesize realistic sequential EHR data; Buczak et al.~\cite{buczak2010data} generate EHR records by utilizing domain-specific knowledge and actual data; Ma et al.~\cite{ma2018risk} proposed the PRIME model to leverage medical knowledge graph for rare medical concepts. However, these existing methods share the following drawback: they do not utilize the large amounts of “non-qualified” EHR records that are excluded during the patient cohort selection process. 

Another approach to tackle the data scarcity problem is transfer learning (also known as model pre-training), which aims to learn good representations in an unsupervised manner to boost model performance in the downstream tasks. Transfer learning has been proven effective in a wide range of Natural Language Processing (NLP) tasks ~\cite{su2009proceedings},~\cite{peters2018deep},~\cite{baevski2019cloze},~\cite{peters2017semi} and Computer Vision (CV) tasks ~\cite{guo2019spottune},~\cite{kan2014domain},~\cite{shao2014transfer}. Recent studies also applied transfer learning techniques to the healthcare domain. For example, Li et al.~\cite{li2020behrt} proposed BEHRT to pre-train the Masked Language Model (MLM) on more than one million patient records to capture the semantic meaning of medical codes, BioBERT~\cite{lee2020biobert} and Clinical-BERT~\cite{alsentzer2019publicly} pre-trained BERT (Bidirectional Encoder Representations from Transformers)~\cite{devlin2018bert} on clinical text for clinical NLP tasks. Despite their promising results, most of these studies focus on pre-training models on free-text; therefore, model pre-training on structural claims data, especially pediatric claims data, are neglected. 

Claims data, a special kind of EHR, which  is mostly used for financial purposes, contains rich clinical information. Such information can reflect the disease progression of patients and offers valuable support for healthcare analysis. Various data-driven models built for different predictive tasks are based on claims data, and, therefore, it makes computational sense to adapt the pre-training framework on claims data. However, pre-training models on claims data have been minimally explored and face unique challenges. First, medical visits are unevenly distributed. The time span between two consecutive medical visits is likely to differ. Second, medical claims contain different types of variables, such as medical codes, claim type, service date, and expenditure. These variables contain clinical meaningful information and need to be modeled collectively Finally, some rare medical codes, such as brain cancer (ICD-9 191.9), have low frequency even in a large healthcare database and, therefore, suffer from severe sparsity.
  
Our goal is to learn a universal representation of claims data that can transfer knowledge with little adaptation to a wide range of population-specific tasks. This study took a semi-supervised approach for predictive healthcare tasks using a combination of unsupervised pre-training and supervised fine-tuning. Specifically, we proposed Claim Pre-Training (Claim-PT), a two-stage framework that first uses Next Visit Prediction (NVP) and Categorial Prediction (CP) as objectives to pre-train the initial parameters of the transformer-based neural network. The NVP predictive objective helps capture the relationships between medical claims and the co-occurrence information of medical codes, while the CP predictive objective induces medical domain knowledge to mitigate the rare medical code sparsity issue. Next, the pre-trained model is adapted to a population-specific predictive task, such as asthma exacerbation prediction, using the corresponding selected patient cohort.

We pre-train our model on a large-scale pediatric claims dataset containing more than one million unique patients. We evaluate our framework on two population-specific predictive tasks (i.e., suicide risk prediction and asthma exacerbation prediction) and compare the model performance with discriminatively trained models. Experimental results show that our model can learn generalized patient representation through encoding the sequential medical claims and significantly outperforms baselines. We also evaluate our model’s transfer learning ability across institutions. The empirical results confirm that our framework possesses great potential for transfer learning across different healthcare organizations, indicating that organizations with insufficient pediatric claims data can also benefit from our pre-trained model.

We publicly release the pre-trained model along with the population-specific preprocessing steps for claims data on GitHub: https://github.com/drxzeng/Claim-PT. In summary, we make the following contributions:

\begin{itemize}
\item We train and publicly release Claim-PT, a transformer-based pre-train \& fine-tune framework trained on a large real-world pediatric claims database. To the best of our knowledge, Claim-PT is the first pre-trained \& fine-tune framework on pediatric claims data, that can deliver a significant performance boost for population-specific predictive tasks.
\item We demonstrate that our framework can utilize the general claims records for medical knowledge understanding. The pre-trained framework helps to improve downstream population-specific medical predictive tasks and outperforms tailored task-specific baselines. In addition, we show that the pre-trained framework has great potential at knowledge generalization across institutions, paving the way for future care coordination and delivery between healthcare organizations. 
\end{itemize}

\section*{Related Work}
In this section, we first review the related research for transfer learning using claims data. Specifically, we focus on deep learning approaches. Next, we present several medical tasks using claims data and the corresponding predictive models.

\subsection*{Transfer Learning using EHR}
Transfer learning is an approach where deep learning models are first trained on a large (unlabeled) dataset to learn generalized parameter initialization and perform similar tasks on another dataset. Several state-of-the-art results in the NLP and CV domain are based on transfer learning solutions~\cite{mikolov2013efficient},~\cite{he2019rethinking},~\cite{radford2018improving}.

Recently, researchers applied transfer learning techniques to the medical domain. Transfer learning enables deep learning models to capture comprehensive contextual semantics, which can benefit the downstream predictive tasks.  For example, MED-BERT~\cite{rasmy2021med} pre-trained contextualized medical code embeddings on large-scale structured EHR and illustrate that the pre-trained embeddings can improve model performance on the downstream tasks. Med2vec, proposed by Choi et al.~\cite{choi2016multi} is a skip-gram-based model that can capture the co-occurrence information between medical visits. Med2vec is able to learn semantic meaningful and interpretable medical code embedding, which can benefit the predictive tasks and provide clinical interpretation. BioBERT~\cite{lee2020biobert}, a pre-trained biomedical language model trained on biomedical text, aims at adapting the language model for biomedical corpora.

These studies demonstrate the effectiveness of the pre-train \& fine-tune framework with respect to boosting model performance on the downstream predictive tasks, especially when the data size is limited. However, none of the previous research focuses on pediatric claims data. We want to explore whether the pre-train \& fine-tune paradigm on pediatric claims can benefit downstream predictive tasks, specifically with a population-specific patient cohort.

\subsection*{Predictive Models using Claims Data}
There has been active research in modeling the longitudinal claims data for various predictive tasks. Generally, these studies can be divided into two groups: works that focus on predicting a specific future medical event, such as suicide risk prediction, asthma exacerbation prediction; and works that focus on a broader range of medical events, such as auto diagnosis and chronic disease progression modeling. 

Various deep learning models have been proposed to model claims data for a specific future medical event prediction. Su et al.~\cite{su2020machine} proposed a logistic regression model with carefully selected features to predict the suicide risk among children. Xiang et al.~\cite{xiang2019asthma} predict the risk of asthma exacerbations and explore the potential risk factors involved in the progression of asthma via a time-sensitive attentive neural network. Zeng et al.~\cite{zeng2021multi} developed a multi-view framework to predict the future medical expenses for better care delivery and care management. Choi et al.~\cite{choi2016retain} proposed RETAIN to estimate the future heart failure rate with explainable risk factors. For general-purpose disease progressing models, Zeng et al.~\cite{zeng2019multilevel} proposed a hierarchical transformer-based deep learning model to forecast future medical events. Ma et al.~\cite{ma2018kame} leverage medical domain knowledge to model the sequential medical codes for the next visit medical code prediction. 

One of the main challenges in developing these models is the size of the dataset. The datasets used in previous studies usually contain over a hundred thousand patients, which is large enough to train most deep learning networks. However, for many population-specific predictive tasks or institutions without a large data corpus, training a complex deep learning model from scratch is not feasible and therefore requires transfer learning or alternative techniques.

\section*{Framework}
In this section, we first describe the motivation of our study under the real-world scenario. Then we describe the formal definition of the problem of pretraining in healthcare using claims data. Finally, we illustrate our proposed methods in detail.

\subsection*{Motivation and Training Procedures}
\begin{figure}[ht]
\centering
\includegraphics[width=0.95\linewidth]{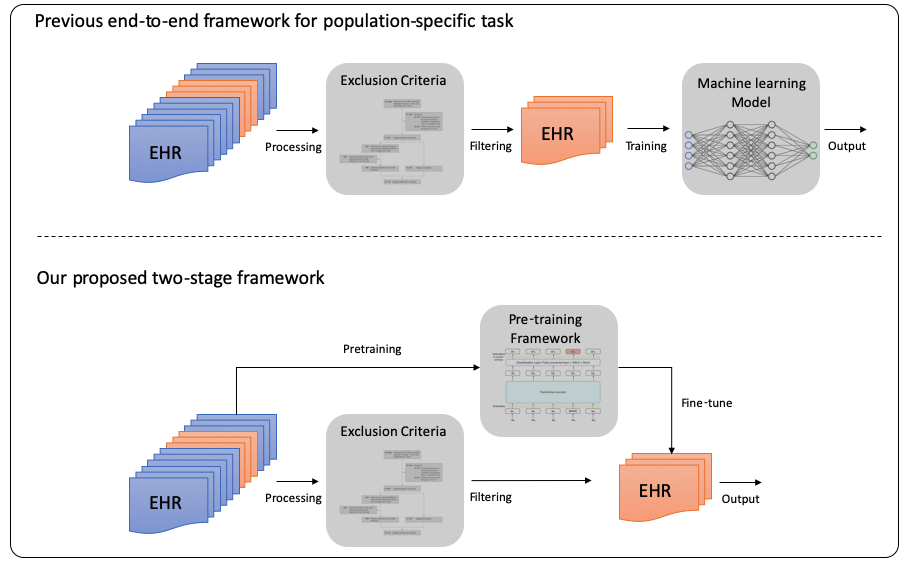}
\caption{The motivation of our study. The traditional pipeline (top) of most medical predictive tasks requires rigorous criteria to construct the case \& control patient cohort, undermining deep learning models' predictive power and leading to suboptimal performance. Our proposed pre-training and fine-tuning framework (bottom) leverage the excluded patient records and significantly boost the model performance.}
\label{fig:fig1}
\end{figure}

Before the formal description of our framework, we would like to explain its underlying motivation and practical scenario by a high-level illustration in Figure \ref{fig:fig1}. Figure \ref{fig:fig1} (top) shows the traditional pipeline of building deep learning models on a predictive task in the medical domain. As one can see, numerous claims records are removed during the cohort selection process following the inclusion and exclusion criteria. After the cohort selection process, only patients with specific diseases or satisfy certain criteria can remain as cases or controls, leaving a relatively small data cohort compared to the original database. The small data size makes it difficult to adequately train a deep learning model and leads to suboptimal model performance.

The observation above provides the main intuition and motivation behind our method. In other words, we aim to leverage the patient records that are excluded in the cohort selection process to boost the model performance on population-specific predictive tasks. We achieve this goal through a two-stage pipeline. The first stage is to learn a high-capacity patient representation learning model, i.e., capturing the clinical meaning of medical codes and the temporal information between medical claims. The first stage is complete through training the model on all available patient records. Next, a fine-tuning stage is followed, where we adapt the pre-trained model to a population-specific medical task. In the situation where the selected patient cohort is small, the pre-trained model can transfer valuable medical information from the excluded patient records.

\subsection*{Unsupervised Pre-training}

Given a patient in the claims data represented as a sequence of medical claims $v_1, v_2, …, vi$ ordered by service date $t$. The i-th medical claim $v_i$ contains a set of medical codes $\{c_1, c_2, …, c_j\}\subseteq|C|$ , where $|C|$ refers to the vocabulary size of the medical codes. There are three different types of medical claims, i.e., inpatient claim $IP$, outpatient claim $OP$, and pharmacy claim $RX$. We use a modified language modeling objective to maximize the likelihood of the medical codes in the next claims, as shown below:
$$L(v_i) =\sum_i log P(v_i | v_1 v_2, …, v_{i-1};\theta)$$
, where $P$ is the conditional probability modeled by a deep neural network with parameters $\theta$. The parameters are trained using Adam optimization algorithm. In addition, we design another objective function aim to mitigate the sparsity issue of rare medical codes:

$$L(v_1, v_2, …, v_{i-1}) =log P( \hat{v}_1, \hat{v}_2,  …, \hat{v}_{i-1} | v_1, v_2, …, v_{i-1};\theta)$$

, where $\hat{v}_i$ contains the categorical medical codes. The second objective guide the model to learn claim representation $v_i$ that are representative of the corresponding categories. Correctly capturing the categories for each medical code within the claims helps to induce expert knowledge and is one of the basis of all downstream prediction tasks.

In our experiments, we adopt the state-of-the-art transformer layer as the building block. The medical codes (i.e., diagnoses, procedures and medications), claims types and service date are first map to latent space via embedding matrices. A max pooling layer is then applied to extract the most salient features within a claim. Next, the transformer block, including a multi-headed self-attention operation, position-wise feedforward layer, and skip-connect normalization layer, is applied to produce an output distribution over the target medical claim. In order to generate patient-level embedding, we also include the demographic information as the first token of every patient sequence. The hidden state of this token is used as the aggregate patient representation for patient-level classification tasks, while the hidden state of the sequential claims hidden state is used for visit-level classification tasks. The equations are shown below :

$$ E_t=[E_{diag};E_{proc};E_{drug};E_{util};E_{date}] $$
$$ e_t=maxpool(E_t) $$
$$pe,\hat{e}_1,\hat{e}_2,…,\hat{e}_t=TransLayer([E_{demo};e_1,e_2,…,e_t]) $$

, where $E_{diag};E_{proc};E_{drug};E_{util};E_{date};E_{demo}$ are the embedding vectors of diagnoses, procedures, medication, claim type, service date and age \& gender. $pe$ is the hidden state for patient embedding and $e_1,e_2,…,e_t$ are the hidden state for claim embeddings.

\subsection*{Population-specific Fine-tuning}

After training the framework with the objective functions as mentioned above, we adapt the parameters to the downstream population-specific predictive task. We assume a population-specific claims dataset that contains both positive subjects (y=1) and negative subjects (y=-1) . The input medical claim sequences and demographic vector are passed through the pre-trained model to obtain the final transformer block’s activation $pe,\hat{e}_1,\hat{e}_2,…,\hat{e}_t$. For patient-level classification task, $pe$ is then fed into an added linear output layer with parameters $W_{pe}$ to predict y:

$$P(y | v_1, v_2, …, v_i)=softmax(pe * W_{pe}).$$

\begin{figure}[ht]
\centering
\includegraphics[width=0.95\linewidth]{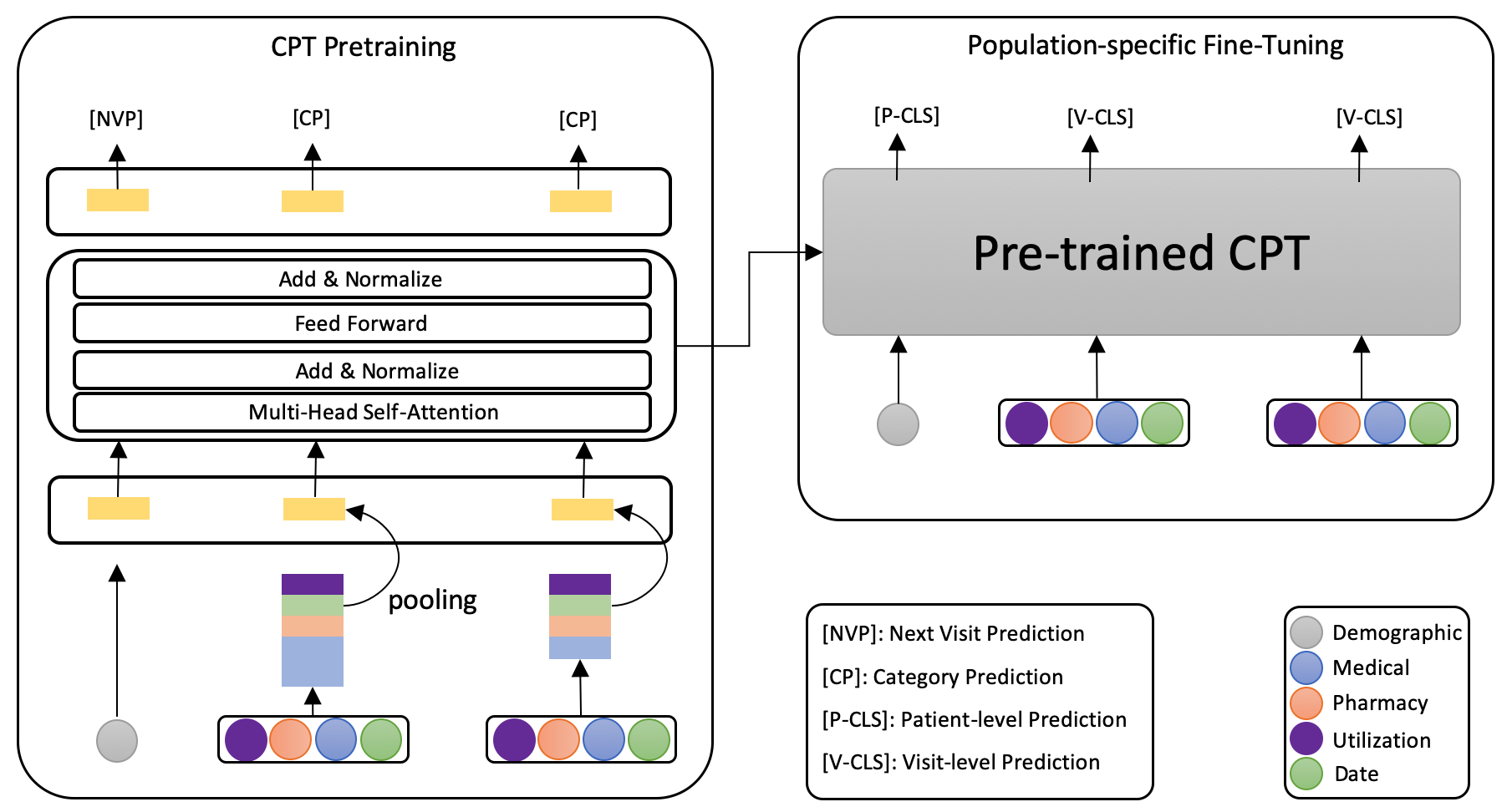}
\caption{The overview of our pretraining \& fine-tuning framework.}
\label{fig:fig2}
\end{figure}

Figure \ref{fig:fig2} illustrate the architecture of our framework. Overall, the only extra parameters we require to train during the fine-tuning stage are $W_{pe}$, and thus we do not need a large patient cohort to train the complex deep learning models from scratch. As a result, our pre-training \& fine-tuning framework are suitable for various population-specific predictive tasks. 

\section*{Results}
This section demonstrates the effectiveness of our Claim-PT framework on different downstream population-specific tasks by comparing it with several baselines. First, we conduct two embedding visualization tasks as sanity checks to validate the model's ability to capture the semantic meaning of patients' medical codes and health conditions. Next, we evaluate the performance of the proposed framework on two real-world medical predictive tasks using claims data. Specifically, we conduct a suicide prediction task and an asthma exacerbation task. The two cases are different in patient-selection criteria and therefore yield different target populations. The criteria and model performance are presented in detail for each of the tasks.

\subsection*{Sanity check: embedding visualization}

Deep learning models understand the data by projecting the inputs into a latent space, where relevant and salient features for solving the problem at hand are well represented and extracted. In our case, the salient features refer to the semantic meaning of patients' medical codes and health conditions. Therefore, we first conduct two sanity checks to verify whether our framework can successfully capture the semantic information of medical codes and stratify patients according to their health conditions.

\begin{figure}[ht]
\centering
\includegraphics[width=0.95\linewidth]{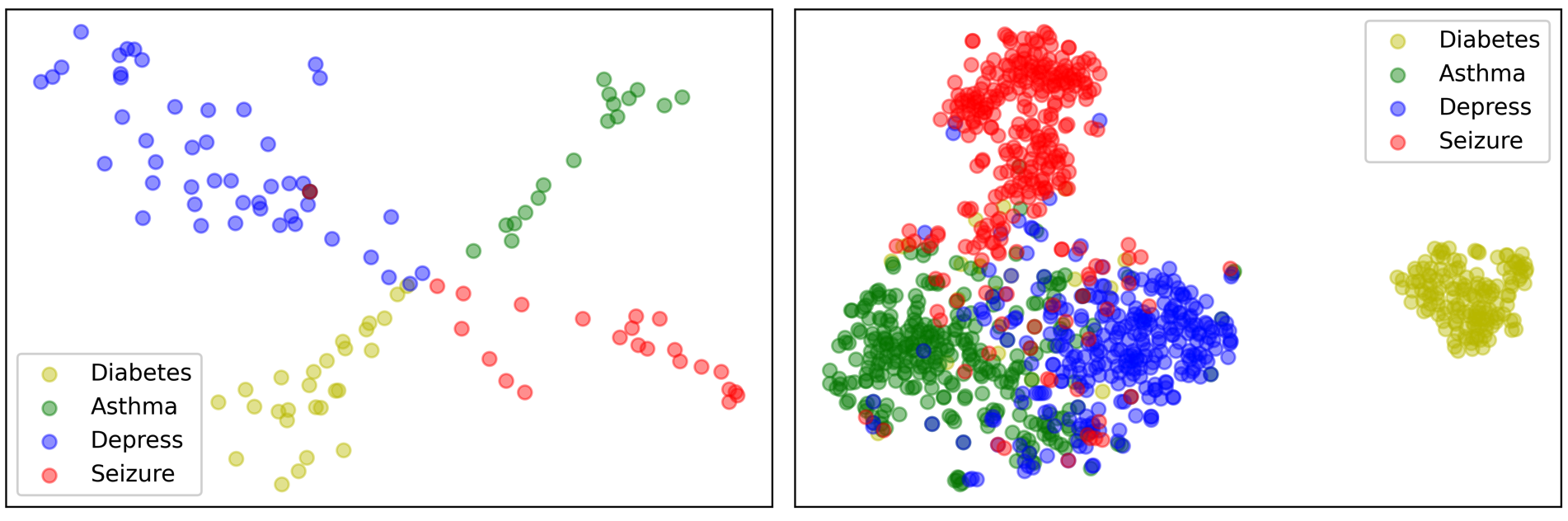}
\caption{ t-SNE scatter plots of diagnosis codes (left) and patients (right) learned by our framework. Four diseases (i.e., diabetes, asthma, depression, and seizure) are selected to provide clear visualization results. ICD-9 diagnosis codes (left) and patients (right) that describe the corresponding disease condition are selected.}
\label{fig:fig3}
\end{figure}

We can observe reassuring patterns from the two embedding visualization plots in Figure \ref{fig:fig3}. For instance, patients with the same disease are grouped together, while patients that are diagnosed with different diseases are well-separated. This observation indicates that our framework is able to capture the clinical meaningful information within claims data and learn efficient embeddings.

\subsection*{Suicide prediction}
Suicide among young people is one of the severe public concerns~\cite{nock2013prevalence}, which is the second leading cause of death for children according to the American Academy of Child and Adolescent Psychiatry (AACAP). In 2017, about 1 in 13 adolescents have attempted suicide, and these suicide attempts cause a significant healthcare burden. Thus, it is critical in clinical practice to predict the risk of suicide accurately. 

Recently, efforts have been made to apply the machine learning architecture to model the sequential claims data to predict suicide risk~\cite{su2009proceedings}. However, the excluding criteria limited the patient population (i.e., only a couple of hundreds of patients have committed suicide in one institution’s database) and therefore restricted the model performance due to inadequately training.

\begin{figure}[ht]
\centering
\includegraphics[width=0.6\linewidth]{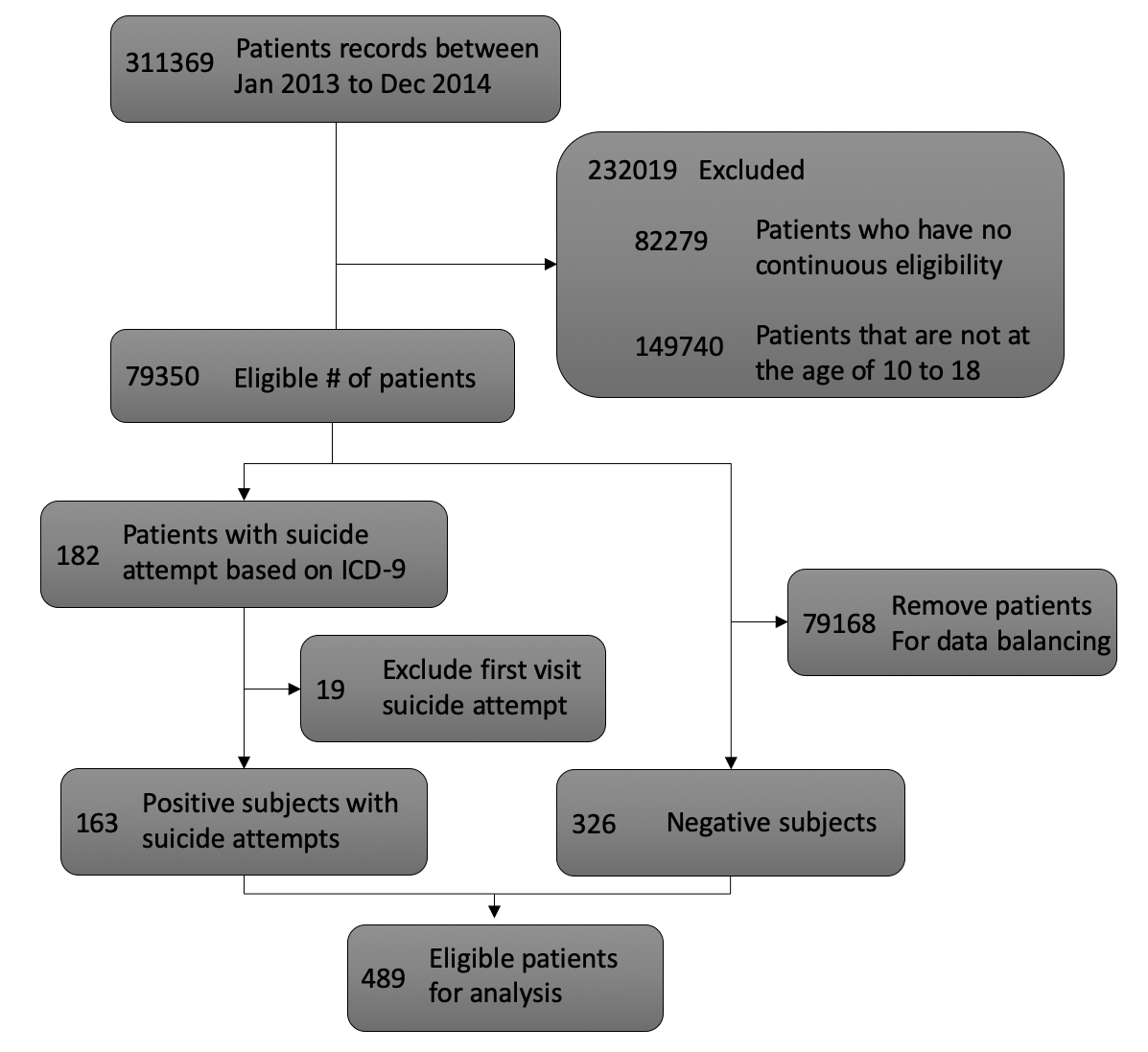}
\caption{The cohort selection process for the study of suicide risk prediction.}
\label{fig:fig4}
\end{figure}

This case study examines whether the excluded patient population can help pre-train the deep learning framework and ultimately provide a model performance boost on suicide risk prediction tasks. In particular, we first construct the patient cohort according to the patient selection criteria shown in Figure \ref{fig:fig4}. 163 patients who have suicide attempts during 2013 to 2014 are selected as positive subjects, while 326 patients who do not have suicide attempts in the same time span are randomly selected as negative subjects. Next, we develop models with the next-visit prediction window, i.e., the model aims to predict the occurrence of a future suicide attempt that happens in the next encounter. Therefore, the model is trained using the patient’s medical visits prior to the first suicide attemp (for positive subjects) or the last recorded medical visit (for negative subjects). All Variables that have been used in the pre-training phase are selected for the suicide prediction task. Finally, we randomly divide the patient cohort into 70\% training and 30\% testing sets. 

All models are trained for twenty epochs to ensure sufficient training. The same experiments are repeated five times, and the average model performances are reported for comparison. We leverage Area Under the receiver operator characteristic Curve (AUC), a widely used evaluation metric of discrimination performance that ranges from 0.5 (random prediction) to 1.0 (perfect prediction), to evaluate the predictive performance for all models. The data preprocessing details can be found on our GitHub page.

From Table~\ref{table:table1}, we can see that the AUC of the proposed pre-training \& fine-tuning framework is higher than that of baselines on the suicide risk prediction task. The Claim-PT variants, Claim-PT Start and Claim-PT Pool, use the pre-trained framework for risk prediction without fine-tuning, and the predicted results achieve similar performance compared to baselines. This indicates that our proposed pre-training objectives is able to capture the semantic meaning of medical codes, and thus the framework can distinguish patients based on their health records. Since the suicide risk prediction task contains less than five hundred patients, deep learning models, including Doctor AI, Dipole, and TransE, can not gain sufficient predictive power and thus yield suboptimal performance compared to linear regression. We can also observe that the model performance between deep learning models is similar. The reason is likely because the number of patient records and the categories of medical codes is limited, and developing more complicated models with more parameters does not help to sufficiently train the parameters.

\begin{table}[ht]
\centering
\begin{tabular}{|l|l|l|}
\hline
 & Model  & AUC \\
\hline
 & LR Sparse~\cite{weisberg2005applied} & 0.73 \\
 & LR Dense~\cite{weisberg2005applied} & 0.74 \\
Cold-start & DoctorAI~\cite{choi2016doctor} & 0.72 \\
 & Dipole~\cite{ma2017dipole} & 0.72 \\
 & TransE~\cite{yuksel2019turkish} & 0.73 \\
\hline
Zero-shot & Claim-PT Start & 0.74 \\
 & Claim-PT Pool & 0.73 \\
\hline
Fine-tune & Claim-PT + Fine-tune & 0.84 \\
\hline
\end{tabular}
\caption{Results on suicide risk prediction task, comparing our framework with baselines via AUC.}
\label{table:table1}
\end{table}

\subsection*{Asthma exacerbation}
We next turn our attention to the asthma exacerbation prediction task. Asthma is a common yet serious health problem among children, costing billions of dollars every year and imposes a significant burden on the healthcare system in the U.S.~\cite{nurmagambetov2018economic}. Asthma exacerbation is a severe asthma condition and requires medical interventions, resulting in an emergency department visit or hospitalization, which is resource-consuming and may even result in death. Therefore, it is practically essential to accurately identify asthma exacerbation in advance for better care delivery and care intervention.

\begin{figure}[ht]
\centering
\includegraphics[width=0.6\linewidth]{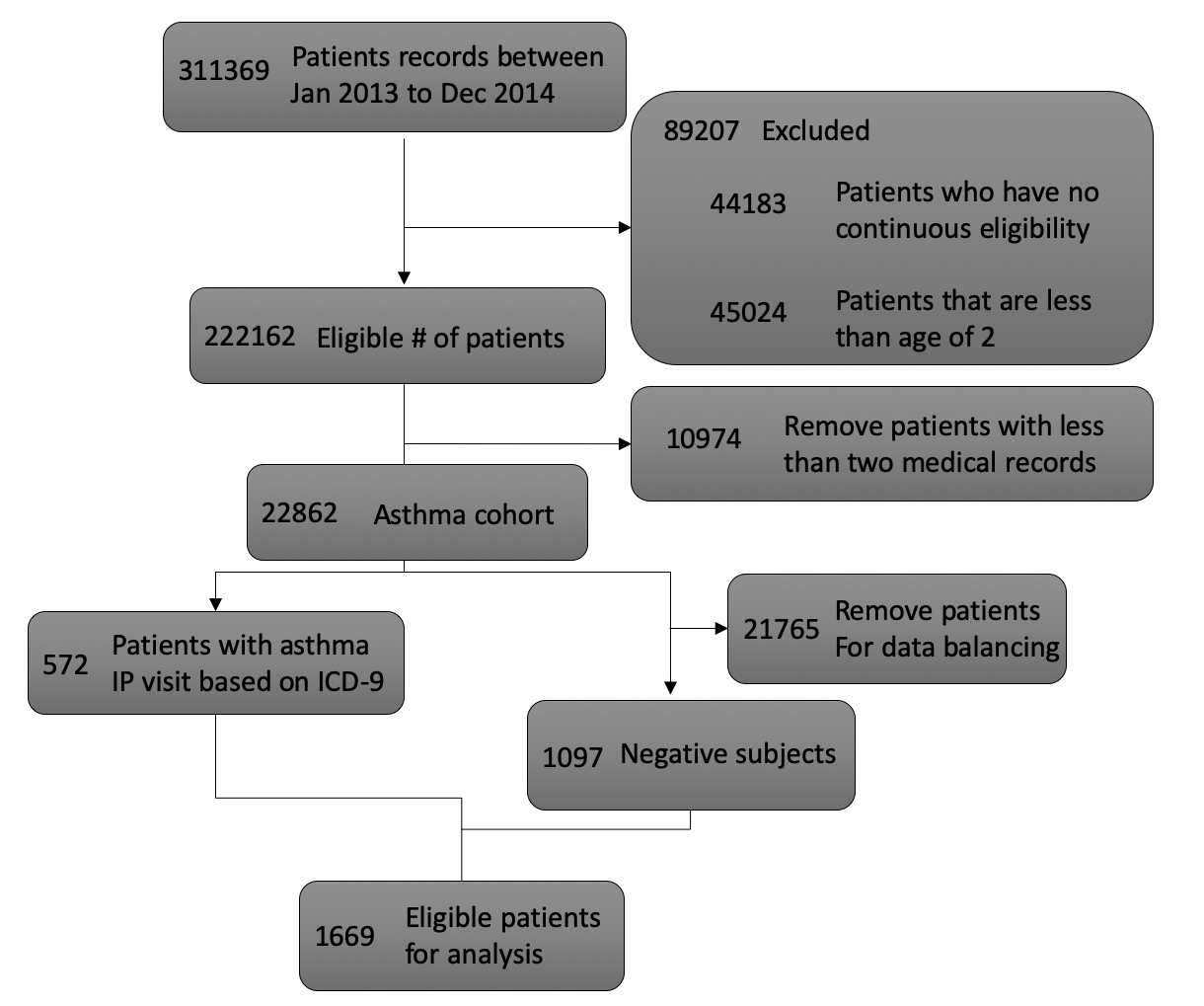}
\caption{The cohort selection process for the study of asthma exacerbation.}
\label{fig:fig5}
\end{figure}

A data-driven study to predict asthma exacerbation using claims data is conducted, and various methods are compared with our Claim-PT framework to illustrate its effectiveness. We adopt the cohort selection process from a recent asthma exacerbation prediction study~\cite{xiang2019asthma}, and the excluding criteria is shown in Figure \ref{fig:fig5}. As shown in the figure, 572 patients with in-patient asthma visits (i.e., patients admitted to hospital due to asthma diagnosis) from 2013 to 2014 are identified as the positive subjects, while 1,097 randomly selected patients without asthma hospitalization are selected as negative subjects. A total of 1,669 patients are identified and form the final cohort. The experiment setup is the same as the suicide risk prediction task, and the detailed preprocessing steps are on our GitHub page.

As shown in Table ~\ref{table:table2}, Our proposed framework showed supervisor model performance compared to all baselines. The AUC of LR Sparse is the highest among all baselines, including the two Claim-PT variants. This observation is likely because the prediction of asthma exacerbation depends on the frequency of specific medical codes, such as the previous hospitalization. LR Sparse captures relationships between these medical codes and the prediction outcome, while LR Dense loses granularity information through feature preprocessing steps. Although RNN-based models, Doctor-AI and Dipole, are able to capture the sequential information, they have a complex model structure and thus yield suboptimal performance. The inferior performance of TransE can confirm this surmise. TransE utilizes the state-of-the-art transformer block as encoder and has more parameters, but the AUC score is the lowest among all approaches.

\begin{table}[ht]
\centering
\begin{tabular}{|l|l|l|}
\hline
 & Model  & AUC \\
\hline
 & LR Sparse~\cite{weisberg2005applied} & 0.77 \\
 & LR Dense~\cite{weisberg2005applied} & 0.73 \\
Cold-start & DoctorAI~\cite{choi2016doctor} & 0.74 \\
 & Dipole~\cite{ma2017dipole} & 0.73 \\
 & TransE~\cite{yuksel2019turkish} & 0.71 \\
\hline
Zero-shot & Claim-PT Start & 0.74 \\
 & Claim-PT Pool & 0.74 \\
\hline
Fine-tune & Claim-PT + Fine-tune & 0.82 \\
\hline
\end{tabular}
\caption{Results on asthma exacerbation prediction task, comparing our framework with baselines via AUC.}
\label{table:table2}
\end{table}

\subsection*{Knowledge transfer across institutes.}
As we observed from the above experiments, the performance of the deep learning model on clinical events is highly dependent on the scale of the dataset. However, many institutions have not yet collected large-scale datasets or can not access sufficient amounts of data due to privacy issues. In these cases, training deep learning from scratch will result in suboptimal model performance and could easily lead to overfitting. One possible way to mitigate this challenge is to transfer knowledge from an institute with a large number of patient records. In this subsection, we evaluate the ability of our framework to transfer knowledge across institutes.

\textbf{PFK-2013}. Ten thousand PFK patients’ claims from Jan 2013 to Dec 2013 are first selected to conduct the experiments. The PFK-2013 is extracted from the PFK dataset (but excluded during the pre-training procedure) to mimic the pediatric claims data from another institution. 

\textbf{MIMIC-3}. A different EHR dataset, MIMIC-3, is used for another across institution experiment. MIMIC-3 is a publicly available clinical dataset that contains ICU patient records for over seven years of observation. This dataset is very different from the PFK pediatric claims data in that it consists of demographically and diagnostically different patients. For example, the median age is 65 in MIMIC-3, while 10 in PFK. The mortality rate among MIMIC-3 patients is about 10\%, while close to zero in PFK. Despite the substantial clinical difference between MIMIC-3 and PFK datasets, they share a similar data structure and therefore suitable for our cross-institution experiment.

\begin{figure}[ht]
\centering
\includegraphics[width=0.95\linewidth]{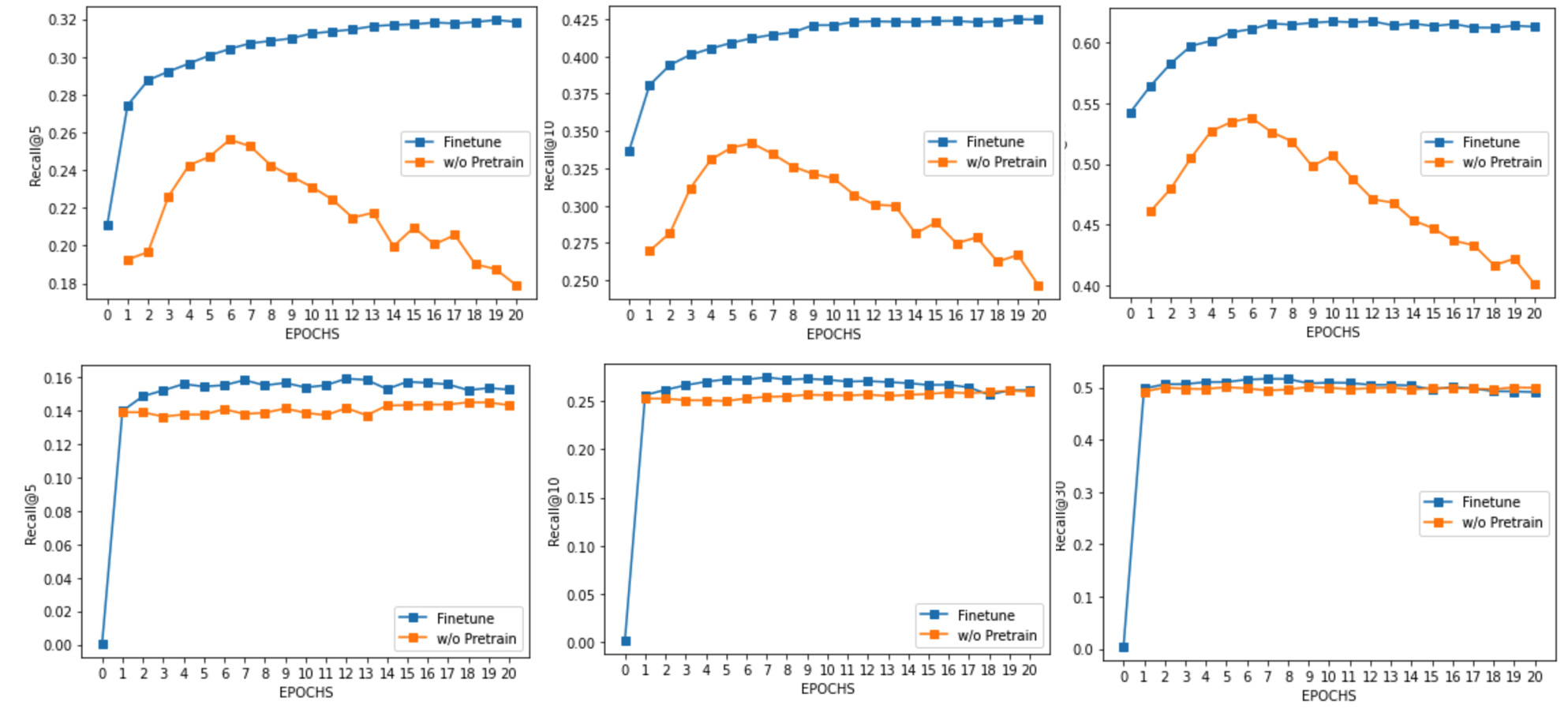}
\caption{The impact of pre-training on improving the performance on PFK-2013 (top) and MIMIC-3 (bottom), PFK-2013 is a subset of the PFK pediatric claims dataset (not used in the pre-training phase), while MIMIC-3 contains ICU patient records. The pre-training procedure results in less than 5\% in the model performance for MIMIC-3, but more than 10\% improvement for PFK-2013.}
\label{fig:fig6}
\end{figure}

Figure\ref{fig:fig6} demonstrates the experiments on PFK-2013 and MIMIC-3. In Figure \ref{fig:fig6} (top), we can observe a vast improvement of the prediction performance induced by knowledge transfer from pre-training. This improvement is because PFK-2013 shares a similar data distribution as PFK (i.e., they both contain pediatric patient records in the Ohio area), enabling the knowledge to transfer from a large dataset to a small dataset successfully. Due to privacy limitations, we can not access another pediatric claims dataset, but we believe the impact of pre-training on improving model performance is still valid. 

In Figure \ref{fig:fig6} (bottom), we test our pre-trained framework on the MIMIC-3 dataset to explore the cross-domain adaptation result. As seen in the figure, the improvement is barely noticeable. The observation matches the clinical intuition as the MIMIC-3 and PFK have a significantly different patient population, where the former contains elderly and severely ill patients in the ICU department and the latter contains children that often visit the outpatient department.

\section*{Conclusion}

This study proposed Claim Pre-Training (Claim-PT), a task-agnostic pre-training \& fine-tuning framework that aims to achieve solid clinical concepts understanding and efficient patient representation learning. By pre-training the transformer-based model on large pediatric claims data with millions of patient records, our model acquires significant medical knowledge, understanding, and ability to process unevenly distributed longitudinal medical claims. The sufficiently pre-trained model can then be leveraged to solve downstream population-specific tasks, which often have a small patient cohort and therefore limit the performance of deep learning models. We conducted three experiments to validate the effectiveness of our Claim-PT framework, i.e., suicide risk prediction task, asthma exacerbation prediction, and auto diagnosis task. The experimental results suggest that our Claim-PT framework can significantly boost model performance on discriminative tasks, achieving more than 10\% performance gain. We also demonstrate that Claim-PT has striking potential to transfer knowledge from one institution to another institution. In summary, our study investigates the possibility of using unsupervised learning to learn general medical knowledge from claims data. The promising results imply that our pre-train \& fine-tune framework can be beneficial in solving downstream population-specific tasks. 

\section*{Data availability}
The data source for this study is claims data from Partner For Kids (PFK). Partner for Kids (PFK) is one of the largest pediatric ACOs for Medicaid enrollees in central and southeastern Ohio. The dataset was obtained from a density sampled study that contains more than 600,000 enrollees’ medical claims. In accordance with the Common Rule (45 CFR 46.102[f]) and the policies of Nationwide Children’s Institutional Review Board, this study used a limited dataset and was not considered human subjects research and thus not subject to institutional review board approval. Restrictions apply to the availability of these data, and so are not publicly available. More information can be found: https://partnersforkids.org/. For help getting access to the dataset, please contact: PartnersForKids@NationwideChildrens.org. 

\bibliography{sample}

@article{choi2017using,
  title={Using recurrent neural network models for early detection of heart failure onset},
  author={Choi, Edward and Schuetz, Andy and Stewart, Walter F and Sun, Jimeng},
  journal={Journal of the American Medical Informatics Association},
  volume={24},
  number={2},
  pages={361--370},
  year={2017},
  publisher={Oxford University Press}
}

@article{landi2020deep,
  title={Deep representation learning of electronic health records to unlock patient stratification at scale},
  author={Landi, Isotta and Glicksberg, Benjamin S and Lee, Hao-Chih and Cherng, Sarah and Landi, Giulia and Danieletto, Matteo and Dudley, Joel T and Furlanello, Cesare and Miotto, Riccardo},
  journal={NPJ digital medicine},
  volume={3},
  number={1},
  pages={1--11},
  year={2020},
  publisher={Nature Publishing Group}
}

@inproceedings{zeng2019multilevel,
  title={Multilevel self-attention model and its use on medical risk prediction},
  author={Zeng, Xianlong and Feng, Yunyi and Moosavinasab, Soheil and Lin, Deborah and Lin, Simon and Liu, Chang},
  booktitle={PACIFIC SYMPOSIUM ON BIOCOMPUTING 2020},
  pages={115--126},
  year={2019},
  organization={World Scientific}
}

@inproceedings{sun2017revisiting,
  title={Revisiting unreasonable effectiveness of data in deep learning era},
  author={Sun, Chen and Shrivastava, Abhinav and Singh, Saurabh and Gupta, Abhinav},
  booktitle={Proceedings of the IEEE international conference on computer vision},
  pages={843--852},
  year={2017}
}

@article{hedderich2018training,
  title={Training a neural network in a low-resource setting on automatically annotated noisy data},
  author={Hedderich, Michael A and Klakow, Dietrich},
  journal={arXiv preprint arXiv:1807.00745},
  year={2018}
}

@article{haines2020testing,
  title={Testing suicide risk prediction algorithms using phone measurements with patients in acute mental health settings: feasibility study},
  author={Haines-Delmont, Alina and Chahal, Gurdit and Bruen, Ashley Jane and Wall, Abbie and Khan, Christina Tara and Sadashiv, Ramesh and Fearnley, David},
  journal={JMIR mHealth and uHealth},
  volume={8},
  number={6},
  pages={e15901},
  year={2020},
  publisher={JMIR Publications Inc., Toronto, Canada}
}

@inproceedings{choi2017generating,
  title={Generating multi-label discrete patient records using generative adversarial networks},
  author={Choi, Edward and Biswal, Siddharth and Malin, Bradley and Duke, Jon and Stewart, Walter F and Sun, Jimeng},
  booktitle={Machine learning for healthcare conference},
  pages={286--305},
  year={2017},
  organization={PMLR}
}

@article{helgheim2019merging,
  title={Merging data diversity of clinical medical records to improve effectiveness},
  author={Helgheim, Berit I and Maia, Rui and Ferreira, Joao C and Martins, Ana Lucia},
  journal={International journal of environmental research and public health},
  volume={16},
  number={5},
  pages={769},
  year={2019},
  publisher={Multidisciplinary Digital Publishing Institute}
}

@inproceedings{seneviratne2018merging,
  title={Merging heterogeneous clinical data to enable knowledge discovery},
  author={Seneviratne, Martin G and Kahn, Michael G and Hernandez-Boussard, Tina},
  booktitle={BIOCOMPUTING 2019: Proceedings of the Pacific Symposium},
  pages={439--443},
  year={2018},
  organization={World Scientific}
}

@article{lee2020generating,
  title={Generating sequential electronic health records using dual adversarial autoencoder},
  author={Lee, Dongha and Yu, Hwanjo and Jiang, Xiaoqian and Rogith, Deevakar and Gudala, Meghana and Tejani, Mubeen and Zhang, Qiuchen and Xiong, Li},
  journal={Journal of the American Medical Informatics Association},
  volume={27},
  number={9},
  pages={1411--1419},
  year={2020},
  publisher={Oxford University Press}
}

@article{buczak2010data,
  title={Data-driven approach for creating synthetic electronic medical records},
  author={Buczak, Anna L and Babin, Steven and Moniz, Linda},
  journal={BMC medical informatics and decision making},
  volume={10},
  number={1},
  pages={1--28},
  year={2010},
  publisher={BioMed Central}
}

@inproceedings{ma2018risk,
  title={Risk prediction on electronic health records with prior medical knowledge},
  author={Ma, Fenglong and Gao, Jing and Suo, Qiuling and You, Quanzeng and Zhou, Jing and Zhang, Aidong},
  booktitle={Proceedings of the 24th ACM SIGKDD International Conference on Knowledge Discovery \& Data Mining},
  pages={1910--1919},
  year={2018}
}

@inproceedings{su2009proceedings,
  title={Proceedings of the Joint Conference of the 47th Annual Meeting of the ACL and the 4th International Joint Conference on Natural Language Processing of the AFNLP},
  author={Su, Keh-Yih and Su, Jian and Wiebe, Janyce and Li, Haizhou},
  booktitle={Proceedings of the Joint Conference of the 47th Annual Meeting of the ACL and the 4th International Joint Conference on Natural Language Processing of the AFNLP},
  year={2009}
}

@article{peters2018deep,
  title={Deep contextualized word representations},
  author={Peters, Matthew E and Neumann, Mark and Iyyer, Mohit and Gardner, Matt and Clark, Christopher and Lee, Kenton and Zettlemoyer, Luke},
  journal={arXiv preprint arXiv:1802.05365},
  year={2018}
}

@article{baevski2019cloze,
  title={Cloze-driven pretraining of self-attention networks},
  author={Baevski, Alexei and Edunov, Sergey and Liu, Yinhan and Zettlemoyer, Luke and Auli, Michael},
  journal={arXiv preprint arXiv:1903.07785},
  year={2019}
}

@article{peters2017semi,
  title={Semi-supervised sequence tagging with bidirectional language models},
  author={Peters, Matthew E and Ammar, Waleed and Bhagavatula, Chandra and Power, Russell},
  journal={arXiv preprint arXiv:1705.00108},
  year={2017}
}

@inproceedings{guo2019spottune,
  title={Spottune: transfer learning through adaptive fine-tuning},
  author={Guo, Yunhui and Shi, Honghui and Kumar, Abhishek and Grauman, Kristen and Rosing, Tajana and Feris, Rogerio},
  booktitle={Proceedings of the IEEE/CVF Conference on Computer Vision and Pattern Recognition},
  pages={4805--4814},
  year={2019}
}

@article{kan2014domain,
  title={Domain adaptation for face recognition: Targetize source domain bridged by common subspace},
  author={Kan, Meina and Wu, Junting and Shan, Shiguang and Chen, Xilin},
  journal={International journal of computer vision},
  volume={109},
  number={1-2},
  pages={94--109},
  year={2014},
  publisher={Springer}
}

@article{shao2014transfer,
  title={Transfer learning for visual categorization: A survey},
  author={Shao, Ling and Zhu, Fan and Li, Xuelong},
  journal={IEEE transactions on neural networks and learning systems},
  volume={26},
  number={5},
  pages={1019--1034},
  year={2014},
  publisher={IEEE}
}

@article{li2020behrt,
  title={BEHRT: transformer for electronic health records},
  author={Li, Yikuan and Rao, Shishir and Solares, Jos{\'e} Roberto Ayala and Hassaine, Abdelaali and Ramakrishnan, Rema and Canoy, Dexter and Zhu, Yajie and Rahimi, Kazem and Salimi-Khorshidi, Gholamreza},
  journal={Scientific reports},
  volume={10},
  number={1},
  pages={1--12},
  year={2020},
  publisher={Nature Publishing Group}
}

@article{alsentzer2019publicly,
  title={Publicly available clinical BERT embeddings},
  author={Alsentzer, Emily and Murphy, John R and Boag, Willie and Weng, Wei-Hung and Jin, Di and Naumann, Tristan and McDermott, Matthew},
  journal={arXiv preprint arXiv:1904.03323},
  year={2019}
}

@article{devlin2018bert,
  title={Bert: Pre-training of deep bidirectional transformers for language understanding},
  author={Devlin, Jacob and Chang, Ming-Wei and Lee, Kenton and Toutanova, Kristina},
  journal={arXiv preprint arXiv:1810.04805},
  year={2018}
}

@article{lee2020biobert,
  title={BioBERT: a pre-trained biomedical language representation model for biomedical text mining},
  author={Lee, Jinhyuk and Yoon, Wonjin and Kim, Sungdong and Kim, Donghyeon and Kim, Sunkyu and So, Chan Ho and Kang, Jaewoo},
  journal={Bioinformatics},
  volume={36},
  number={4},
  pages={1234--1240},
  year={2020},
  publisher={Oxford University Press}
}

@article{mikolov2013efficient,
  title={Efficient estimation of word representations in vector space},
  author={Mikolov, Tomas and Chen, Kai and Corrado, Greg and Dean, Jeffrey},
  journal={arXiv preprint arXiv:1301.3781},
  year={2013}
}

@inproceedings{he2019rethinking,
  title={Rethinking imagenet pre-training},
  author={He, Kaiming and Girshick, Ross and Doll{\'a}r, Piotr},
  booktitle={Proceedings of the IEEE/CVF International Conference on Computer Vision},
  pages={4918--4927},
  year={2019}
}

@article{radford2018improving,
  title={Improving language understanding by generative pre-training},
  author={Radford, Alec and Narasimhan, Karthik and Salimans, Tim and Sutskever, Ilya},
  year={2018}
}

@article{rasmy2021med,
  title={Med-BERT: pretrained contextualized embeddings on large-scale structured electronic health records for disease prediction},
  author={Rasmy, Laila and Xiang, Yang and Xie, Ziqian and Tao, Cui and Zhi, Degui},
  journal={NPJ Digital Medicine},
  volume={4},
  number={1},
  pages={1--13},
  year={2021},
  publisher={Nature Publishing Group}
}

@inproceedings{choi2016multi,
  title={Multi-layer representation learning for medical concepts},
  author={Choi, Edward and Bahadori, Mohammad Taha and Searles, Elizabeth and Coffey, Catherine and Thompson, Michael and Bost, James and Tejedor-Sojo, Javier and Sun, Jimeng},
  booktitle={Proceedings of the 22nd ACM SIGKDD International Conference on Knowledge Discovery and Data Mining},
  pages={1495--1504},
  year={2016}
}

@article{su2020machine,
  title={Machine learning for suicide risk prediction in children and adolescents with electronic health records},
  author={Su, Chang and Aseltine, Robert and Doshi, Riddhi and Chen, Kun and Rogers, Steven C and Wang, Fei},
  journal={Translational psychiatry},
  volume={10},
  number={1},
  pages={1--10},
  year={2020},
  publisher={Nature Publishing Group}
}

@article{xiang2019asthma,
  title={Asthma Exacerbation Prediction and Interpretation based on Time-sensitive Attentive Neural Network: A Retrospective Cohort Study},
  author={Xiang, Yang and Ji, Hangyu and Zhou, Yujia and Li, Fang and Du, Jingcheng and Laila, Rasmy and Wu, Stephen T and Zheng, Wenjin Jim and Xu, Hua and Zhi, Degui and others},
  journal={medRxiv},
  pages={19012161},
  year={2019},
  publisher={Cold Spring Harbor Laboratory Press}
}

@article{choi2016retain,
  title={Retain: An interpretable predictive model for healthcare using reverse time attention mechanism},
  author={Choi, Edward and Bahadori, Mohammad Taha and Kulas, Joshua A and Schuetz, Andy and Stewart, Walter F and Sun, Jimeng},
  journal={arXiv preprint arXiv:1608.05745},
  year={2016}
}

@inproceedings{choi2016doctor,
  title={Doctor ai: Predicting clinical events via recurrent neural networks},
  author={Choi, Edward and Bahadori, Mohammad Taha and Schuetz, Andy and Stewart, Walter F and Sun, Jimeng},
  booktitle={Machine learning for healthcare conference},
  pages={301--318},
  year={2016},
  organization={PMLR}
}

@inproceedings{ma2018kame,
  title={Kame: Knowledge-based attention model for diagnosis prediction in healthcare},
  author={Ma, Fenglong and You, Quanzeng and Xiao, Houping and Chitta, Radha and Zhou, Jing and Gao, Jing},
  booktitle={Proceedings of the 27th ACM International Conference on Information and Knowledge Management},
  pages={743--752},
  year={2018}
}

@article{nock2013prevalence,
  title={Prevalence, correlates, and treatment of lifetime suicidal behavior among adolescents: results from the National Comorbidity Survey Replication Adolescent Supplement},
  author={Nock, Matthew K and Green, Jennifer Greif and Hwang, Irving and McLaughlin, Katie A and Sampson, Nancy A and Zaslavsky, Alan M and Kessler, Ronald C},
  journal={JAMA psychiatry},
  volume={70},
  number={3},
  pages={300--310},
  year={2013},
  publisher={American Medical Association}
}

@article{nurmagambetov2018economic,
  title={The economic burden of asthma in the United States, 2008--2013},
  author={Nurmagambetov, Tursynbek and Kuwahara, Robin and Garbe, Paul},
  journal={Annals of the American Thoracic Society},
  volume={15},
  number={3},
  pages={348--356},
  year={2018},
  publisher={American Thoracic Society}
}

@inproceedings{ma2017dipole,
  title={Dipole: Diagnosis prediction in healthcare via attention-based bidirectional recurrent neural networks},
  author={Ma, Fenglong and Chitta, Radha and Zhou, Jing and You, Quanzeng and Sun, Tong and Gao, Jing},
  booktitle={Proceedings of the 23rd ACM SIGKDD international conference on knowledge discovery and data mining},
  pages={1903--1911},
  year={2017}
}

@inproceedings{yuksel2019turkish,
  title={Turkish tweet classification with transformer encoder},
  author={Y{\"u}ksel, At{\i}f Emre and T{\"u}rkmen, Ya{\c{s}}ar Alim and {\"O}zg{\"u}r, Arzucan and Alt{\i}nel, Berna},
  booktitle={Proceedings of the International Conference on Recent Advances in Natural Language Processing (RANLP 2019)},
  pages={1380--1387},
  year={2019}
}

@book{weisberg2005applied,
  title={Applied linear regression},
  author={Weisberg, Sanford},
  volume={528},
  year={2005},
  publisher={John Wiley \& Sons}
}

\end{document}